\documentclass[lettersize,journal]{IEEEtran}
\usepackage{amsmath,amsfonts}
\usepackage{algorithmic}
\usepackage{algorithm}
\usepackage{array}
\usepackage[caption=false,font=normalsize,labelfont=sf,textfont=sf]{subfig}
\usepackage{textcomp}
\usepackage{stfloats}
\usepackage{url}
\usepackage{verbatim}
\usepackage{graphicx}
\usepackage{cite}
\usepackage[justification=centering]{caption}
\usepackage{hyperref}
\usepackage{orcidlink}

\usepackage{tabularx}   
\usepackage{array}      
\usepackage{booktabs}   
\usepackage{multirow}

\usepackage[dvipsnames]{xcolor}
\begin{document}

\title{Polar Perspectives: Evaluating 2-D LiDAR Projections for Robust Place Recognition with Visual Foundation Models}







\author{Pierpaolo Serio\,\orcidlink{0009-0002-5569-4591‬}, Giulio Pisaneschi\,\orcidlink{https://orcid.org/0000-0001-7873-6008‬},Andrea Dan Ryals\,\orcidlink{https://orcid.org/0000-0001-7173-8406‬}, Vincenzo Infantino\,\orcidlink{https://orcid.org/0000-0001-7873-6008‬},\\ Lorenzo Gentilini\,\orcidlink{https://orcid.org/0000-0001-7873-6008‬}, Valentina Donzella\,\orcidlink{https://orcid.org/0000-0002-3408-6135‬}, Lorenzo Pollini\,\orcidlink{https://orcid.org/0000-0002-9165-2594‬}

\thanks{The source codes will be publicly available after publication }}



\maketitle

\begin{abstract}
This work presents a systematic investigation into how alternative LiDAR–to–image projections affect metric place recognition when coupled with a state-of-the-art vision foundation model. We introduce a modular retrieval pipeline that controls for backbone, aggregation, and evaluation protocol, thereby isolating the influence of the 2-D projection itself. Using consistent geometric and structural channels across multiple datasets and deployment scenarios, we identify the projection characteristics that most strongly determine discriminative power, robustness to environmental variation, and suitability for real-time autonomy. Experiments with different datasets, including integration into an operational place recognition policy, validate the practical relevance of these findings and demonstrate that carefully designed projections can serve as an effective surrogate for end-to-end 3-D learning in LiDAR place recognition.

\end{abstract}


\section{Introduction}

Robotics and Artificial intelligence (AI) form a deeply interconnected technological field in which software intelligence is embedded into machines capable of sensing, moving, and acting in the physical world. AI techniques, ranging from classical control and planning to modern machine learning and large-scale neural networks, enable robots to interpret sensor data, understand their surroundings, and make context-aware decisions. At the same time, advances in mechatronics, sensors, edge computing, and low-latency communication have dramatically broadened where and how robots can operate. Once limited to structured factory floors, robots now assist in medical procedures, inspect infrastructure, collaborate with humans in warehouses, and explore hazardous or remote environments.
A particularly influential domain within this broader field is autonomous systems. These include autonomous ground vehicles such as self-driving cars and delivery robots, aerial systems like drones, and marine platforms ranging from surface vessels to deep-sea explorers and also space robotics \cite{lunarhopper}. Robotic vehicles rely heavily on perception (e.g., computer vision, lidar processing, sensor fusion), simultaneous localization and mapping (SLAM), predictive modeling of other agents, and motion-planning algorithms that ensure safe, efficient navigation. Within this domain, reliable long-term localization and mapping remain fundamental challenges for autonomous agents operating in real-world environments\cite{sousa2023systematic}. In SLAM pipelines, small, systematic pose errors accumulate over time, producing drift that corrupts the map and degrades downstream tasks such as planning and looped mission execution\cite{guclu2019fast}. Detecting loop closures is therefore a keystone capability for accurate, robust mapping and long-term autonomy. This process is conventionally decomposed into two sequential steps: Place Recognition, which involves correctly identifying that the sensor platform has returned to a previously visited location, and Pose Estimation, which utilizes this recognition to compute a relative transformation and enforce global pose consistency through optimization. Over the last decade, advances in deep learning have transformed how perception systems approach this problem: learned descriptors and metric-learning objectives produce compact embeddings that can be indexed for large-scale retrieval, and self-supervised transformer backbones, now often referred to as visual foundation models, yield dense, transferable patch-level features that generalize across scenes and domains. For robotics, these developments are double-edged: they enable high-recall, data-driven proposals for loop closure and simplify cross-modal transfer, yet they also introduce concerns about real-time performance, domain shift, interpretability, and the need to integrate learned outputs with provably robust geometric estimation. Consequently, state-of-the-art place recognition systems increasingly combine learning-based retrieval (fast, scalable, and semantically informed) with classical geometric verification to meet the strict reliability, latency, and safety requirements of deployed robotic platforms. The richest and most informative sensor modality for grasping accurate geometrical information of places in outdoor robotics is arguably the LiDAR point cloud\cite{li2020lidar,zhou2023lidar, Vergara2024Complementing}: an unstructured set of 3-D points that directly encodes scene geometry at sensor resolution, is largely invariant to illumination, and provides the metric information needed for precise localization. Nevertheless, processing an entire point cloud end-to-end with deep models for place recognition remains an open challenge until very recently because there were no widely available foundation models that produce dense, transferable descriptors from raw 3-D scans at the scale and quality needed for robust retrieval. In this work we take a complementary approach: rather than learning from raw point clouds directly, we ask whether and how one can best convert a point cloud into a 2-D representation that preserves the cues essential for distinguishing places and expresses them in a form that current visual foundation models can process effectively. Briefly, our research question is: what 2-D projection of a LiDAR scan yields the most discriminative and robust place descriptor when processed by a modern visual backbone? To answer this we evaluate a single, high-quality retrieval pipeline built on a vision foundation model encoder coupled with an aggregation model across several point cloud–to-image mappings (bird’s-eye view, front-view / range image, and other engineered projections). This empirical comparison isolates the effect of the 2-D representation itself and shows how best to exploit existing vision foundation models for metric place recognition with LiDAR data.

\subsection{Contributions}
Within this scope, the present work offers the following contributions:
\begin{itemize}
    \item We introduce a modular pipeline built upon the state-of-the-art DINOv3 Vision Foundation Model (VFM)\cite{dinov3}, enabling a systematic comparison of arbitrary 2-D projections of LiDAR point clouds. The framework accommodates projections of any spatial resolution and/or channel configuration and supports benchmarking across four diverse datasets, including both intra-sequence and inter-sequence place-recognition scenarios. The entire system is released publicly to facilitate reproducibility and to support further research by the community.
    \item Through controlled experiments that fix the retrieval backbone and aggregation strategy, we isolate the influence of the 2-D projection itself. This yields to the identification of which projection characteristics most strongly determine discriminative performance and robustness under different deployment conditions and environments.
    \item Each examined representation is enriched with a consistent set of channels—range or height, local curvature, and sensor intensity—providing both geometric details and mid-level regional structure information.
    \item We evaluate the proposed approach in a challenging unstructured warehouse environment characterized by dynamic objects, and substantial viewpoint variation. These experiments include a real loop-closure policy to assess the practical utility of the learned descriptors in realistic SLAM settings, further demonstrating the relevance of the findings for long-term autonomy.
\end{itemize}

\section{Related works}

As LiDAR Place Recognition (LPR) plays a fundamental role within Loop Closure for Simultaneous Mapping and Localization\cite{zhang20243d}, enormous efforts are made to solve this problem, starting with traditional Computer Vision techniques and then moving to the rise of Machine Learning ones. 
Due to the high number of points that charactherize LiDAR pointclouds, the modern approaches aim to tackle the Place Recognition problem in an holistic way by finding a global descriptor that grasp the fundamental and invariant features of a pointcloud\cite{vidanapathirana2022logg3d,khaliq2019holistic}. Using the raw 3D pointcloud, M2DP introduces the multiview 2D projection (M2DP), a global point cloud descriptor. It projects 3D points onto 2D planes to capture spatial density distributions. The descriptor is subsequently constructed from the singular vectors of these distributions\cite{he2016m2dp}.
PointNetVLAD was proposed as a deep network for large-scale 3D point cloud retrieval to enable place recognition. This architecture integrates PointNet and NetVLAD for end-to-end extraction of a global descriptor\cite{uy2018pointnetvlad}. The pioneering Scan Context (SC) method introduced a non-histogram-based global descriptor that captures the structural appearance of a 3D environment by partitioning the horizontal space and retaining maximum point height, forming a 2D matrix\cite{kim2018scan}. It achieves achieved its invariance to LiDAR viewpoint changes  through its descriptor structure and an efficient two-phase search algorithm utilizing a similarity score (ring key and column-wise comparison). While demonstrating promising performance, particularly in rotation invariance, the original SC may exhibit reduced efficacy under significant lateral offsets. Subsequent research has led to numerous SC-based variants that incorporate features such as polar/Cartesian context, intensity, and frequency domain information to enhance the descriptor's overall robustness\cite{kim2021scan,wang2020intensity}. The Radon Sinogram (RING) framework and its successor, RING++, represent other pivotal approaches for LPR intended for use with sparcity and under conditions of large viewpoint differences.  The initial RING method introduced a compact, unified representation that is theoretically proven to exhibit orientation- and translation-invariance, with a certifiable robustness against large pose differences between the query and map scans\cite{lu2022one}. The latter RING++ methodology relies on processing a Bird's-Eye View (BEV) of the scan, exploiting the mathematical properties of the Radon and Fourier Transforms to generate a roto-translation-invariant place representation that keeps the certifiability\cite{xu2023ring++}. To mitigate the computational burden and inherent complexities associated with processing raw, dense point clouds, certain research efforts adopt the Bird's Eye View (BEV) projection as a streamlined point cloud representation. This representation inherently facilitates the fusion of LiDAR measurements with camera imagery within a multiview setting, thereby harnessing the complementary strengths of both sensor modalities for enhanced perception and robustness\cite{wang2018fusing,luo2021bvmatch}. BEVPlace and BEVPlace++ constitute a set of LPR methodologies that leverage the BEV representation to establish robustness against view variations and sensor motion effects. The initial BEVPlace utilizes group convolution in conjunction with NetVLAD for the generation of a rotation-invariant global feature, which capitalizes on the BEV's observed stability under sensor translation\cite{luo2023bevplace}. BEVPlace++ extends this approach by integrating a Rotation Equivariant Module (REM) within a REIN architecture, which is designed to enhance robustness to view changes. This structure facilitates the production of rotation-equivariant local features and rotation-invariant global descriptors, trained solely with coarse place labels for supervision\cite{luo2025bevplace++}. The generalization and pattern recognition capabilities of vision foundational models (VFM) have rapidly driven research, initially focusing on camera imagery before quickly extending to LiDAR point cloud analysis\cite{keetha2023anyloc,mirjalili2023fm,awais2025foundation,qiu2024emvp}.
Attempts to employ Visual Foundation Models (VFMs) directly on point clouds have been made\cite{puy2024three,wu2025sonata}; however, these approaches are constrained to object detection or registration tasks and maintain a reliance on visual images. To achieve this, DINOv2\cite{oquab2023dinov2} and its own features have been leveraged from surround-view images as cross-modal point descriptors for scan-to-map localization\cite{vodisch2025lidar,keetha2023anyloc,huang2024dino}. The ImLPR pipeline presents a LiDAR Place Recognition (LPR) method that adapts DINOv2 \cite{jung2025imlpr}. The system utilizes a specialized Range-Image View (RIV) that encodes the point cloud using three channels: reflectivity, range, and normal ratio. This RIV input allows the RGB-pretrained VFM to extract features via lightweight convolutional adapters inserted into the network architecture, preserving the majority of DINOv2's pre-trained weights. The training process incorporates the Patch-InfoNCE loss, a patch-level contrastive function, to enhance the local discriminability and robustness of the learned LiDAR features. This structure aims to combine LiDAR's geometric accuracy with the VFM's representational capacity.



\begin{figure*}[ht]
    \centering \includegraphics[width=\textwidth]{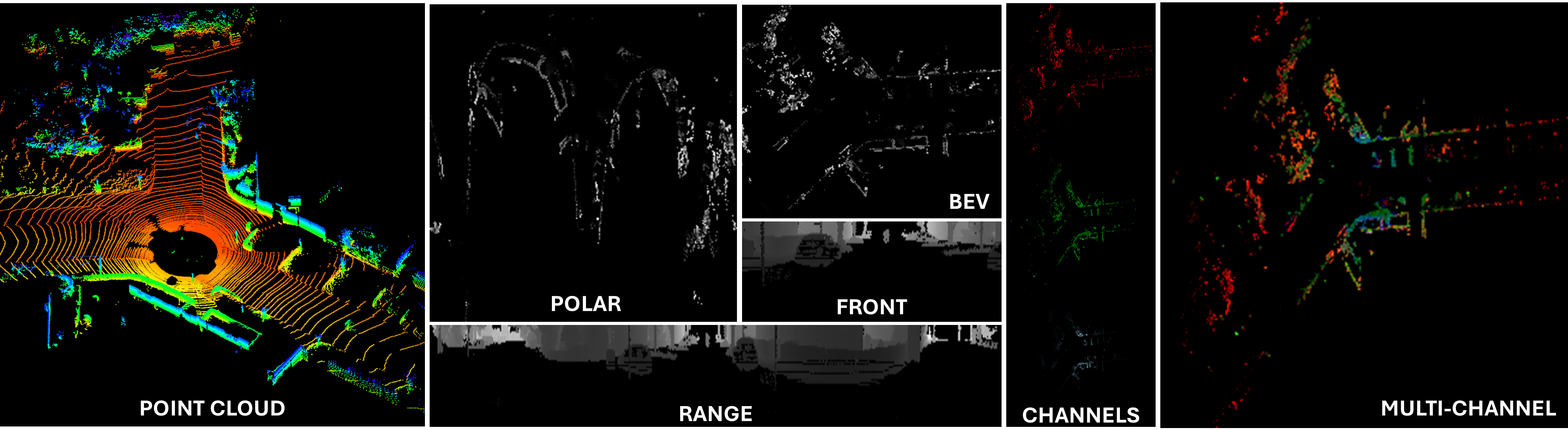}
    \caption{Example of different representations obtained from a single pointcloud, channel information extraction and multi-channel image.}
    \label{fig:represention}
\end{figure*}

\section{Methods}
\label{sec:Methods}
\subsection{Problem definition}

As the vehicle traverses an environment, we treat each acquired sensor frame \(z_t\) (a LiDAR scan converted to a 2-D representation) at time \(t\) and spatial location \(l_t\) as a \emph{place}. The map \(\mathcal{M}\) is the ordered collection of frames observed up to the current time;

In our experiments, we extract two sets of frames: the \emph{database}, which is the set from which we search for matches, and \emph{queries}, where we look for correspondences with a temporal offset to avoid near-duplicates.

In our experiments, we extract a \emph{database} (frames for which we search for matches) and a set of \emph{queries} (frames in which we look for correspondences, with a temporal offset to avoid near-duplicates).

Revisitedness is assessed only for temporally separated observations.

We model the description function by a learned encoder \(f_\theta\) that maps each input \(z\) to a compact global descriptor
\[
g = \xi(z)\in\mathbb{R}^d,
\]

Revisitedness (ground-truth positive relation) is defined geometrically from pose data: a database frame \(j\) is a positive for query \(q\) iff the Euclidean distance between their positions satisfies
\[
\|p_q-p_j\|_2 < \tau.
\]
An additional constraint is introduced to prevent trivial self-matches: enforcing a minimum temporal separation \(|t_q-t_j|>\Delta t\) ensures that the algorithm does not accidentally compare a frame with itself or with an immediately adjacent frame, which would artificially reduce the error and bias the estimation. 
Positive matches that also satisfy the temporal condition are defined as \emph{place recognitions}.

 \subsection{Representation and retrieval pipeline}\label{subsection:representation}


We treat place recognition as a descriptor-based retrieval task.
Each input frame $\mathcal{I}_q \in \mathbb{R}^{H\times W \times 3}$ is mapped through a mapping function $\xi = f \circ \alpha$ in a latent embedding space as a global descriptor $g_q \in \mathbb{R}^{L}$.
This operation consists of two steps. First, the frame is mapped in a feature space as a dense set of downsampled maps $\mathbf{F}$  through $f: \mathbb{R}^{H\times W \times 3} \xrightarrow{} \mathbb{R}^{c\times h \times w} $ with $h<H$ and $w<W$. These features are then compressed into a global descriptor $\mathbf{g}_q$ through an aggregator module $\alpha: \mathbb{R}^{c\times h \times w} \xrightarrow{} \mathbb{R}^{L} $ with $L << H\cdot W$.
This compression allows to decrease the memory burden of storing each image during the navigation phase. The network backbone is a Vision Foundation Model: a self-supervised, transformer-based architecture that processes the input image as a grid of patches and produces robust, patch-level feature tokens. These spatial tokens are converted into a single, fixed-length place descriptor by a projection head (a pooling layer that aggregates spatial information into a global vector). At test time we \(L_2\)-normalize the global vectors,
\[
\tilde g_i = \frac{g_i}{\|g_i\|_2},
\]
and retrieve candidates by nearest-neighbour search in this latent space,
\[
\hat j=\arg\min_{j\in\mathcal D}\|\tilde g_q-\tilde g_j\|_2,
\]
which is equivalent to ranking by cosine similarity because for unit vectors \(\|\tilde g_q-\tilde g_j\|_2^2 = 2 - 2\,\tilde g_q^\top \tilde g_j\). Normalization therefore, makes the retrieval criterion insensitive to vector magnitude and focuses the comparison on direction, which empirically improves robustness to global feature scaling.

To align the learned descriptors with the place-recognition objective, we fine-tune the model using a gold-standard triplet training routine\cite{schroff2015facenet,park2024evaluation}. Concretely, training uses pose-aware sampling: temporally or spatially nearby frames are sampled as positives (they represent the same place under different views), while distant frames are sampled as negatives (different places). This sampling strategy teaches the network to pull together embeddings of the same place seen from different viewpoints and to push apart embeddings of different places. We train the pretrained transformer backbone with a smaller learning rate than the projection head, so that task-specific adjustments concentrate in the head, while the backbone preserves and slightly adapts its general-purpose representations; this separation stabilises fine-tuning and speeds convergence toward descriptors that are robust to viewpoint change and small translations. We evaluate retrieval with Recall@1 — the fraction of queries that have at least one ground-truth positive in the database and for which the top-1 retrieved item is a positive. To characterise score–threshold trade-offs we record the top-1 distance distribution and compute a precision–recall curve by thresholding those distances.

\subsection{Model Structure}
As discussed in \ref{subsection:representation} our model for place recognition is made up of a backbone that extract feature maps from the input and a projection head that embeds the feature map in a single dimensional vector. 
\paragraph{Backbone}

The model backbone is made up from Dinov3\cite{dinov3}, a state-of-the-art Vision Foundation Model.
From an AI/ML perspective, DINOv3 is trained by large-scale self-supervised learning. 
A principal stabilization introduced for DINOv3, commonly referred to as \emph{Gram anchoring}, encourages preservation of the token–token correlation structure across different views during long training runs. Intuitively, this constrains the model to maintain consistent relations among tokens so that dense features do not collapse to trivial solutions and remain useful for matching as well as for global summarization. Since such a large model requires significant resources to run, we employ in our framework a smaller, distilled variant of it, that is "ViT-B/16" (which we call DinoV3 for sake of simplicity), with a reduction from 7 billion to 84 million parameters. Given an input image \(I\), DINOv3 produces a sequence of token embeddings $\mathbf{F} = \{f_i (I)\}, \,\ f_i (I)\in \mathbb{R}^{c\times h \times w}$ where $c = 768$, $h=H/16$, and $w = W/16$. These tokens combine semantic richness with spatial locality: they capture high-level scene semantics while remaining discriminative at patch scale, which is essential when the same place must be recognized across viewpoint and appearance change. Because of its scale, training regimen and empirical robustness, DINOv3 functions effectively as a general-purpose pretrained encoder whose token embeddings transfer well to many downstream tasks. In our pipeline we use these token embeddings as the primary feature representation: they are either pooled to form a single place descriptor or left intact as dense inputs to subsequent pooling. Practically, DINOv3 can be used frozen (as a stable feature provider) or fine-tuned with a small learning rate when task-aligned adaptation is desired.

\paragraph{Projection Heads}
To evaluate the robustness of Dino's feature maps, we employ two different projection heads. As a first solution, we employ Mean + Standard Deviation Pooling, a simple projection that captures first and second order statistics. Given the downsampled feature map $\textbf{F} \in \mathbb{R}^{c\times h \times w}$
the mean pooling for the $c^{th}$ feature map is computed as:
\begin{equation}
    \mathbf{\mu}_c = \frac{1}{h \cdot w} \sum_{i=1}^{h} \sum_{j=1}^{w} \mathbf{F}_{c,i,j} \,.
\end{equation} 
This pooling embeds information about the dominant patterns and common features in the scene. 
The standard deviation pooling instead, grasps information about the distribution spread and texture complexity. This pooling is computed as:
\begin{equation}
    \mathbf{\sigma}_c = \sqrt{\frac{1}{h \cdot w} \sum_{i=1}^{h} \sum_{j=1}^{w} (\mathbf{F}_{c,i,j} - \mathbf{\mu}_c)^2}
    \,.
\end{equation}
The resulting global descriptor $\mathbf{g}$ is obtained by stacking the result of the two poolings as $\mathbf{g} = [\boldsymbol{\mu}; \boldsymbol{\sigma}] \in \mathbb{R}^{2c}$.
The second projection head uses NetVLAD\cite{arandjelovic2016netvlad}. NetVLAD is a learned aggregation mechanism designed to convert a set of spatial token embeddings into a single fixed-length vector that summarizes a place. It operates by soft-assigning each token to a set of \(K\) learnable cluster centers \(\{c_k\}_{k=1}^K\) and aggregating residuals relative to those centers. Denoting the tokens by \(\{f_t\}_{t=1}^T\) and the soft-assignment of token \(t\) to center \(k\) by \(a_k(f_t)\), the per-cluster aggregated residual is
\[
g_k \;=\; \sum_{t=1}^T a_k(f_t)\,\bigl(f_t - c_k\bigr)\qquad (k=1,\dots,K),
\]
and the unnormalized global descriptor is the concatenation \(v=[v_1,\dots,v_K]\). A short projection and normalization pipeline (involving dimensionality reduction, whitening and \(L_2\)-normalization in practice) produces the final compact global descriptor \(g\) used for retrieval. This pooling scheme has two important advantages for place recognition. First, the residual aggregation explicitly encodes how the current frame’s local structure deviates from learned visual words, producing a descriptor sensitive to distinctive place cues rather than only to average content. Second, because pooling is learned end-to-end, cluster centers and assignment functions adapt to the retrieval objective during training with pose-aware supervision, yielding global vectors that emphasize features discriminative of geographic proximity rather than generic object semantics.

\subsection{Point-cloud preprocessing and 2-D representations}

We preprocess each LiDAR scan and convert it to several two-dimensional representations suitable for a visual retrieval pipeline. Let a raw scan be the point set $P=\{p_i\}_{i=1}^N$ with $p_i=(x_i,y_i,z_i,I_i)$. We first apply spatial filters to limit the region of interest and removing the road floor. We compute a local curvature measure for each point and append it as an extra channel; the processed points are therefore $(x,y,z,i,\kappa)$.

Each 2-D representation maps points to integer pixel coordinates $(u,v)$ on an $H\times W$ grid and stores normalized per-point values in one or more channels. The four projections used are:

\subsubsection{Bird’s-eye view (BEV)} The Bird’s-eye view projection provides a top-down, metrically view of the scene and is conceptually the closest 2-D analogue of the original 3-D geometry. Because LiDAR sensors primarily capture the horizontal organization of the environment—roads, building facades, vegetation boundaries, and parked vehicles—compressing the point cloud onto the ground plane preserves a large fraction of the spatial cues that are discriminative for place recognition. To generate this representation, points are first recentered to guarantee positive coordinates within the grid:
\[
x'_i=x_i-\min_j x_j,\quad y'_i=y_i-\min_j y_j,
\]

after which they are linearly mapped to discrete pixel indices according to the spatial extent of the scan:

\[
u_i=\left\lfloor \dfrac{x'_i}{\max_j x'_j}(H-1)\right\rfloor,\quad
v_i=\left\lfloor \dfrac{y'_i}{\max_j y'_j}(W-1)\right\rfloor.
\]

Each point deposits one or more normalized quantities—typically height $z$ for the single-channel BEV, or height, intensity, and curvature for the multi-channel variant—into the corresponding pixel. In this form, large-scale geometric layout becomes easily interpretable by a 2-D visual encoder: height variations reflect local terrain and object structure, while intensity and curvature add local surface cues that help disambiguate places with similar global layout. Because BEV images maintain the metric relationships of the original ground plane, they often serve as a strong baseline for LiDAR-to-image place recognition.

\subsubsection{Polar Projection}
The polar representation reorganizes the point cloud in terms of radial distance and planar angle, aligning naturally with the sensor's radial acquisition pattern. Each point is mapped using
\begin{align*}
&v_i = \left\lfloor \frac{r_i}{\max_j r_j}\,(H-1)\right\rfloor,\\
&u_i = \left\lfloor \frac{\theta_i - \min_j \theta_j}{\max_j\theta_j - \min_j\theta_j}\,(W-1)\right\rfloor.
\end{align*}
where $r_i=\sqrt{x_i^2+y_i^2}$ and $\theta_i=\operatorname{atan2}(y_i,x_i)$. This projection emphasizes radial structures and introduces a degree of rotation robustness, which can be advantageous when viewpoint changes occur across traversals. As in the Cartesian case, available per-point features are normalized and stored as pixel values.

\subsubsection{Range Image Projection}

The range image preserves the LiDAR's intrinsic beam sampling pattern by mapping each point according to its azimuth $\theta$, elevation $\phi$, and range $r$:
\begin{align*}
&r_i = \sqrt{x_i^2+y_i^2+z_i^2},\\
&\theta_i=\operatorname{atan2}(y_i,x_i),\\
&\varphi_i=\operatorname{asin}(z_i/r_i).
\end{align*}
Pixels are obtained via
\[
u_i=\left\lfloor 0.5\left(1-\frac{\theta_i}{\pi}\right)\,W\right\rfloor,\qquad
v_i = \arg\min_k |\varphi_i - \gamma_k|,
\]
where $\Gamma = \left[\gamma_1, .. ,\gamma_n\right]$  denotes the LiDAR's factory-calibrated vertical beam angles. For each pixel, the minimum range among all points mapping to that pixel is stored, mirroring the visibility properties of true range-view LiDAR imagery. This geometry-aware projection preserves fine-grained structural cues closely aligned with the physical sensor model.

\subsubsection{Front-View Projection}
The front-view projection constructs a camera-like cylindrical depth image by restricting the horizontal field of view to $[a_{\min}, a_{\max}]$. Azimuth and elevation are computed as above, and image indices follow
\[
u_i = \left\lfloor \frac{\theta_i - a_{\min}}{a_{\max}-a_{\min}}\,W \right\rfloor,\qquad
v_i = \arg\min_k |\varphi_i - \gamma_k|.
\]
Each pixel stores the smallest observed range, producing a depth-like representation that remains faithful to LiDAR geometry while naturally matching the 2-D grid structure expected by visual encoders. This view emphasizes front-facing geometry, which is particularly relevant in autonomous driving scenarios.

One example for each of the proposed projections is depicted in Figure \ref{fig:represention}. 
Across all projections, we apply per-channel normalization to $[0,1]$ (range channel divided by a chosen maximum range) and resize outputs to a fixed $H\times W$ for the visual encoder. When multiple points collide into the same pixel the range image retains the closest point (min reduction); other channels use the last assignment (alternatively, aggregation schemes may be used). These diverse projections (BEV, polar, range, front-view), evaluated under the same descriptor-based retrieval pipeline, allow us to identify which 2-D encoding of a LiDAR scan best preserves place-relevant structure for retrieval.

\subsection{Channel definitions and interpretation}
The pipeline represents each LiDAR point with a small set of per-point scalars that are propagated into image channels. In the code, a point is stored as
\[
p_i = (x_i,\,y_i,\,z_i,\,I_i,\,\kappa_i),
\]
where \(x,y,z\) are Cartesian coordinates, \(I\) denotes intensity, and \(\kappa\) denotes an estimated local curvature. The couple $(x_i,y_i)$ is processed during the point to point conversion between different representations, while the other variables are accordingly adjusted to properly encode the channel related information, with appropriate normalization applied before composing the image.

Two complementary per-point channels used throughout the representations are \emph{intensity} and \emph{curvature}. Below we summarize what they represent, how they are computed in our pipeline, and practical notes on normalization and limitations.

\paragraph*{Intensity} The intensity \(I_i\) is the LiDAR return amplitude associated with point \(p_i\). Physically, intensity depends on the surface reflectivity, incidence angle, and range. In our pipeline the raw intensity value is read from the sensor record and treated as a scalar per point that provides a photometric-like cue complementary to geometry. 

\paragraph*{Curvature} Curvature is an estimate of local geometric variation around a point and is computed from the eigenstructure of the covariance of a point's \(K\)-nearest neighbours. Let \(\mathcal{N}_K(p_i)\) be the set of \(K\) neighbours around \(p_i\). The \(3\times 3\) covariance matrix is
\begin{align*}
C_i \;=\; \frac{1}{K}&\sum_{p_j\in\mathcal{N}_K(p_i)} (p_j-\mu_i)(p_j-\mu_i)^\top,
\\
\qquad &\mu_i=\frac{1}{K}\sum_{p_j\in\mathcal{N}_K(p_i)} p_j.
\end{align*}
Let \(\lambda_{1}\ge\lambda_{2}\ge\lambda_{3}\ge 0\) be the eigenvalues of \(C_i\). A robust measure of local surface variation is curvature, defined as:
\[
\kappa_i \;=\; \frac{\lambda_{3}}{\lambda_{1}+\lambda_{2}+\lambda_{3}},
\]
which ranges near zero on planar patches and increases on edges and corners. In our preprocessing, the scalar \(\kappa_i\) is appended to each point and then min--max normalized per-frame before projection into image channels. 

\subsection*{Per-point channels and their use in 2-D projections}

In the bird’s eye view (BEV) image each pixel stores the normalized triplet \([z,I,\kappa]\) for the point mapped to that cell so that height encodes vertical geometry while intensity and curvature provide complementary material and local-shape cues. In the polar representation, projection points are converted to \((r,\theta)\) with \(r=\sqrt{x^2+y^2}\) and \(\theta=\operatorname{atan2}(y,x)\), discretized into radial and angular bins, and the remaining channels \((z,I,\kappa,\ldots)\) are stored at the corresponding bin after normalization. In the range-image projection, we compute per-point range and elevation
\[
r_i=\sqrt{x_i^2+y_i^2+z_i^2},\qquad \varphi_i=\arcsin(z_i/r_i),
\]
map azimuth \(\operatorname{atan2}(y_i,x_i)\) linearly to columns via
\[
u_i=\left\lfloor 0.5\Bigl(1-\frac{\operatorname{atan2}(y_i,x_i)}{\pi}\Bigr)W\right\rfloor,
\]
and assign rows to the closest physical LiDAR beam using the elevation table \(\{\alpha_k\}\),
\[
v_i=\arg\min_k\bigl|\mathrm{deg}(\varphi_i)-\alpha_k\bigr|.
\]
The image channels store the normalized selected features (range, intensity, curvature) written from the mapped points. The front-view image is analogous to the range image but restricted to a horizontal field-of-view \([a_{\min},a_{\max}]\). Intensity encodes return amplitude (material/reflectivity and incidence effects) and curvature \(\kappa\) is computed from the eigenvalues of the local neighbor covariance (e.g., \(\kappa=\lambda_3/(\lambda_1+\lambda_2+\lambda_3)\)), so intensity gives photometric-like cues and curvature encodes local geometric distinctiveness; both are complementary but sensitive to sampling density, occlusion and sensor noise and therefore treated as auxiliary, normalized channels rather than absolute measures.

\section{Dataset and Evaluation Criteria}
In this section, we describe the datasets and evaluation
criteria that we employ for performance validation.

 \subsection{Dataset}
 The proposed architecture has been tested on three common datasets for place recognition tasks and on a dataset collected from an unstructured warehouse with dynamic elements. This combination was chosen to exercise complementary challenges: KITTI stresses loop-closure and high-quality outdoor LiDAR scans; NCLT exposes long-term, seasonal and appearance changes; HELILPR evaluates sensor heterogeneity and rotationally-rich urban geometry; and our WAREHOUSE sequence probes dynamic, cluttered, and indoor–outdoor conditions that occur in real deployments.
 \paragraph*{KITTI}\cite{Geiger2013IJRR}: This dataset provides 64-ray LiDAR scans (from a Velodyne HDL-64E). The sequences 02, 05, and 06 exhibit a high incidence of loop closure events, making them established benchmarks for evaluating Simultaneous Localization and Mapping (SLAM) and loop closure algorithms. We tested the intra-sequence place recognition and retrieval performances by splitting LiDAR measurements into a query split and a database split as explained in Section \ref{subsection:Testing},  following common practice in recent LiDAR place-recognition works\cite{cui2022bow3d,cattaneo2022lcdnet,luo2025bevplace++}. 
 
\begin{figure}[ht!]
    \centering
    \includegraphics[width=\columnwidth]{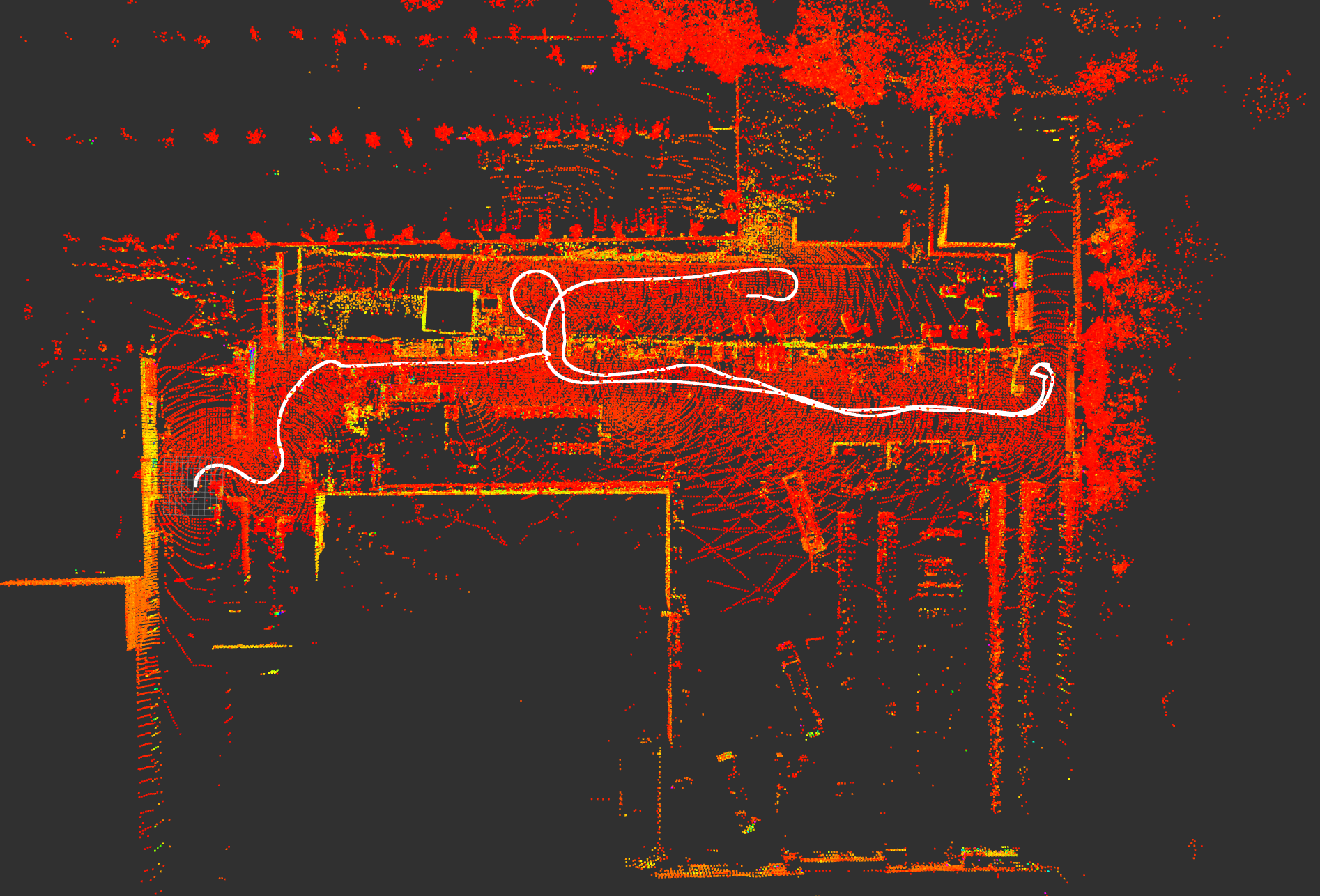}
    \caption{Trajectory followed by the vehicle in the Warehouse environment}
    \label{fig:warehouse}
\end{figure}

\begin{figure}[ht!]
     \centering
\includegraphics[width=0.4\linewidth]{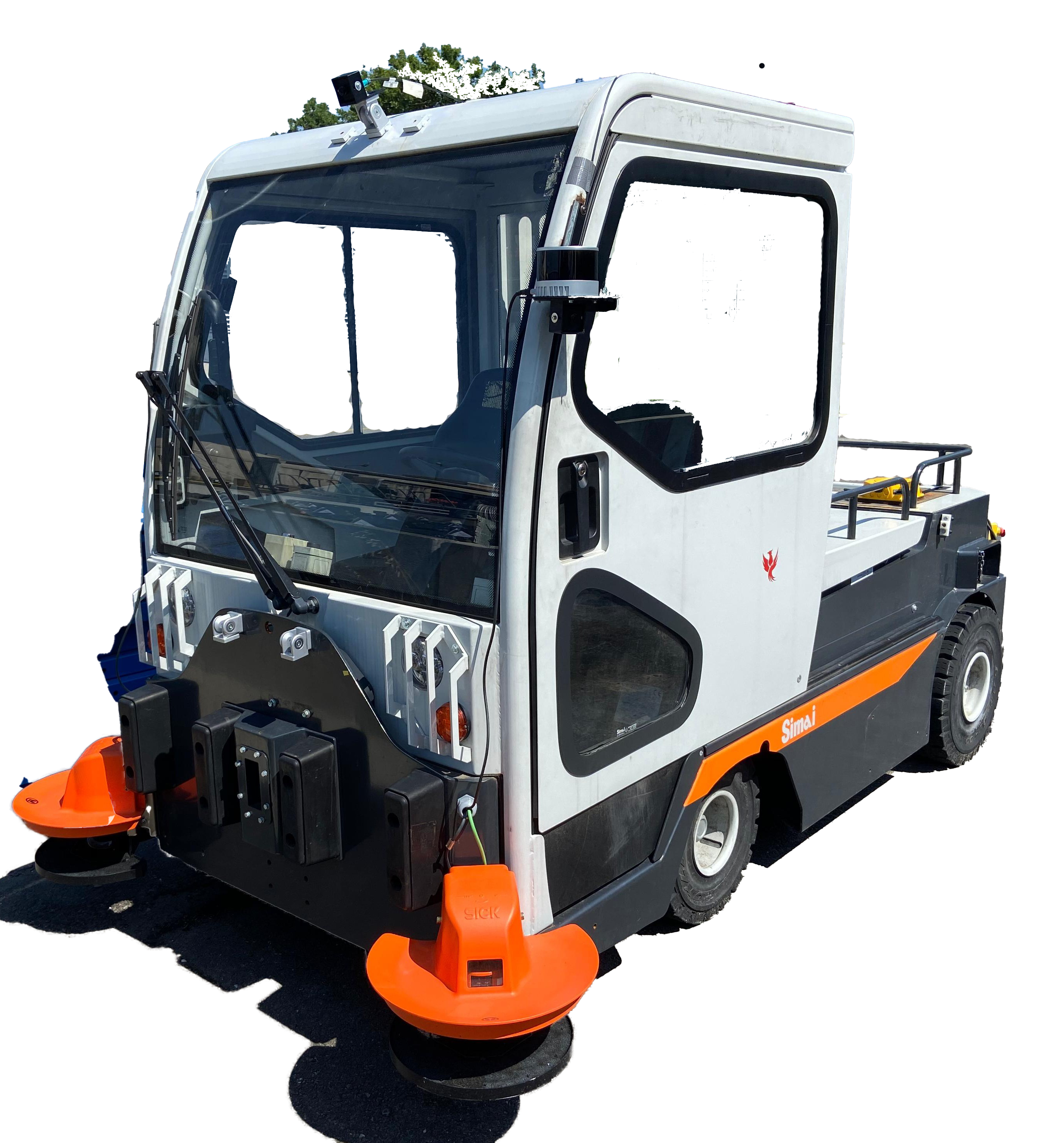}
     \caption{The towing vehicle used in the Warehouse environment}
     \label{fig:towing}
 \end{figure}

 \paragraph*{NCLT}: The North Campus Long-Term (NCLT) dataset contains multiple traversals of the same places collected over months and years, making it ideal for assessing long-term, inter-sequence place recognition under seasonal and viewpoint variation\cite{ncarlevaris2015a}. Because of its design, we employ this dataset to evaluate the inter-sequences place-recognition capabilities. In particular, the scene $2012-01-15$ as database and the scenes $2012-02-04, 2012-06-15, 2013-02-23$ as queries, using the data collected by the Velodyne HDL-32E.
 \paragraph*{HELILPR}: The Heterogeneous LiDAR Dataset for inter-LiDAR Place Recognition (HELILPR) dataset is well known for its diverse urban environments and multiple traversal sessions\cite{jung2024helipr}. the ROUNDABOUT sequence’s large circular layout and outer hexagon road pattern create many revisitable viewpoints and pronounced rotational variation, which helps test descriptor invariance to rotation and repeated geometry that is uncommon in straight urban roads.
 We use "ROUNDABOUT" sequences 01 through 03, specifically employing the data captured by the Velodyne VLP-16 spinning LiDAR with an intra-sequence database-query split.
 \paragraph*{WAREHOUSE}: In addition to all the datasets widely used in literature, we run our model on a custom-made dataset. This dataset presents challenging conditions including indoor-outdoor transitions and dynamic elements such as pedestrians and vehicles. The trajectory exhibits multiple viewpoint variations and significant changes in scene geometry, ranging from confined indoor spaces with limited field of view to open outdoor areas with extended visibility ranges. Furthermore, the sequence includes substantial illumination variations and reflective surfaces typical of industrial environments, which pose additional challenges for LiDAR-based perception systems.
The data acquisition vehicle is depicted in Figure \ref{fig:towing} and is equipped with a Hesai XT32M LiDAR sensor. Ground truth poses are obtained by executing the DLO (Direct LiDAR Odometry) algorithm~\cite{chen2022direct}. An example trajectory followed by the vehicle in a Wharehouse environment is displayed in Figure\ref{fig:warehouse}.

 \subsection{Evaluation Metrics}

We report three complementary metrics that together quantify retrieval accuracy, operating-point behaviour, and confidence calibration.  First, a single-number accuracy metric (Recall@1) captures whether the system retrieves a correct match at the top rank; second, a threshold-based analysis of the top-1 distances produces a precision–recall curve and a \(\max\)-\(F_1\) score, which identify useful operating points when a distance threshold is used as a rejection rule; third, the PR AUC summarises the overall precision–recall trade-off and is robust to class imbalance. 

The primary scalar metric reported is Recall@1 (R@1):
\[
\text{R@1} \;=\; \frac{\#\{q\in\mathcal Q:\ \text{positives}(q)\neq\emptyset\ \wedge\ \hat j_q\in\text{positives}(q)\}}{\#\{q\in\mathcal Q:\ \text{positives}(q)\neq\emptyset\}}.
\]
This measures top-1 retrieval accuracy on the subset of queries that actually have at least one ground-truth match in the database.  R@1 is a direct and interpretable indicator of retrieval success: in many robotics and mapping applications a single correct top-ranked retrieval is sufficient to trigger downstream actions (e.g., loop closure), so a high R@1 directly reflects useful system behaviour.

To study the precision–recall trade-off we convert top-1 distances into binary accept/reject decisions by thresholding. Let \(D=\{d_q\}\) be the distances for queries that have at least one positive and \(L=\{\ell_q\}\) be the corresponding binary labels (\(\ell_q=1\) if \(\hat j_q\) is a true positive, else \(0\)). For a threshold \(t\) the predicted positives are
\[
\hat \ell_q(t)=\mathbf{1}[d_q < t].
\]
We choose the convention that smaller distances indicate higher confidence in a match, so a query is ``predicted positive'' when its top-1 distance falls below the threshold.

Precision and recall are computed in the usual way:
\[
\text{Precision}(t)=\frac{\sum_q \hat \ell_q(t)\,\ell_q}{\sum_q \hat \ell_q(t)},\qquad
\text{Recall}(t)=\frac{\sum_q \hat \ell_q(t)\,\ell_q}{\sum_q \ell_q},
\]
where the product $\hat{l}_q(t),l_q$ are the True Positives (TP), $\hat{l}_q(t)$ does not distinguish between a True Positive (TP) and a False Positive (FP) and $l_q(t)$ does not distinguish between a True Positives (TP) and a False Negatives (FN). Hence, the well-known it is possible to express precision and recall @ distance $t$ using the standard formulations:
\begin{equation}
\mathrm{Precision}(t) = \frac{TP}{TP + FP},
\qquad
\mathrm{Recall}(t)    = \frac{TP}{TP + FN},
\label{eq:prec_recall}
\end{equation}

Scanning \(t\) across the observed interval \([\min D, \max D]\) yields a precision–recall curve. The \(\mathbf{max\text{-}F_1}\) score is defined as follows:
\begin{equation}
    F_1 = 2 \,\times\, \frac{\text{Precision}\times\text{Recall}}{\text{Precision}+\text{Recall}}.
\end{equation}
\(\max\)-\(F_1\) identifies the best single threshold that balances precision and recall, which is useful when a deployment requires a single, fixed distance cutoff.

The area under the precision–recall curve (PR AUC) is obtained by sorting the sampled points by recall and integrating precision with respect to recall using the trapezoidal rule. PR AUC summarises performance across all thresholds: unlike a single operating point, it reflects the overall confidence calibration of the top-1 distances and is particularly informative when the class balance (number of positives vs. negatives) is skewed.

We therefore report: (i) \(\text{Recall@1}\) (percent) to quantify top-1 accuracy on relevant queries; (ii) \(\max\)-\(F_1\) to indicate the best achievable balance between precision and recall under a distance threshold; and (iii) PR AUC to summarise the global precision–recall behaviour of the top-1 distances as a confidence measure.

\begin{table*}[htbp]
\caption{Results of the models for the KITTI, NCLT, and HELILPR datasets. For each sequence, R, MF1, and AUC are reported.}
\centering

\begin{tabular}{l@{\hspace{5em}}ccc@{\hspace{5em}}ccc@{\hspace{5em}}ccc}
\toprule
\multicolumn{10}{c}{\textbf{KITTI}} \\
\midrule
\textbf{Method}
& \multicolumn{3}{c@{\hspace{6em}}}{\textbf{02}} & 
\multicolumn{3}{c@{\hspace{6em}}}{\textbf{05}} & 
\multicolumn{3}{c@{\hspace{2em}}}{\textbf{06}} \\
\midrule
& \textbf{R@1} & \textbf{F1} & \textbf{AUC}
& \textbf{R@1} & \textbf{F1} & \textbf{AUC}
& \textbf{R@1} & \textbf{F1} & \textbf{AUC} \\
\midrule
\multicolumn{10}{c}{\textbf{Netvlad}} \\
\midrule
BEV   & 0.912 & 0.954 & 0.967 & 0.949 & 0.981 & 0.995 & 0.970 & 0.980 & 0.979 \\
Polar & 0.927 & 0.956 & 0.985 & 0.967 & 0.986 & 0.997 & 1.00 & 0.995 & 0.985 \\
Range & 0.804 & 0.974 & 0.973 & 0.895 & 0.958 & 0.994 & 1.00 & 0.995 & 0.985 \\
Front & 0.709 & 0.915 & 0.897 & 0.842 & 0.924 & 0.983 & 0.98 & 0.995 & 0.993 \\
\midrule
\multicolumn{10}{c}{\textbf{MS}} \\
\midrule
BEV   & 0.785 & 0.954 & 0.964 & 0.900 & 0.966 & 0.991 & 0.970 & 0.989 & 0.992\\
Polar & 0.855 & 0.964 & 0.976 & 0.953 & 0.979 & 0.995 & 1.00 & 0.995 & 0.985 \\
Range & 0.794 & 0.970 & 0.948 & 0.901 & 0.968 & 0.995 & 1.00 & 0.995 & 0.985\\
Front & 0.701 & 0.906 & 0.887 & 0.850 & 0.932 & 0.977 & 0.970 & 0.984 & 0.979\\
\bottomrule
\end{tabular}
\vspace{1.2em}

\begin{tabular}{l@{\hspace{5em}}ccc@{\hspace{5em}}ccc@{\hspace{5em}}ccc}
\toprule
\multicolumn{10}{c}{\textbf{NCLT}} \\
\midrule
\textbf{Method}
& \multicolumn{3}{c@{\hspace{6em}}}{\textbf{2012-02-04}} & 
\multicolumn{3}{c@{\hspace{6em}}}{\textbf{2012-06-15}} & 
\multicolumn{3}{c@{\hspace{2em}}}{\textbf{2013-02-23}} \\
\midrule
& \textbf{R@1} & \textbf{F1} & \textbf{AUC}
& \textbf{R@1} & \textbf{F1} & \textbf{AUC}
& \textbf{R@1} & \textbf{F1} & \textbf{AUC} \\
\midrule
\multicolumn{10}{c}{\textbf{Netvlad}} \\
\midrule
BEV   & 0.776 & 0.876 & 0.862 & 0.744 & 0.853 & 0.813 & 0.524 & 0.697 & 0.623 \\
Polar & 0.779 & 0.876 & 0.868 & 0.761 & 0.867 & 0.874 & 0.520 & 0.689 & 0.652 \\
Range & 0.584 & 0.792 & 0.816 & 0.554 & 0.749 & 0.767 & 0.145 & 0.336 & 0.240 \\
Front & 0.444 & 0.660 & 0.552 & 0.371 & 0.546 & 0.426 & 0.105 & 0.215 & 0.117 \\
\midrule
\multicolumn{10}{c}{\textbf{MS}} \\
\midrule
BEV   & 0.634 & 0.838 & 0.856 & 0.714 & 0.841 & 0.845 & 0.425 & 0.693 & 0.667 \\
Polar & 0.675 & 0.874 & 0.919 & 0.769 & 0.889 & 0.932 & 0.490 & 0.739 & 0.763 \\
Range & 0.514 & 0.729 & 0.748 & 0.479 & 0.674 & 0.679 & 0.135 & 0.287 & 0.196 \\
Front & 0.433 & 0.643 & 0.548 & 0.343 & 0.515 & 0.408 & 0.117 & 0.225 & 0.118 \\
\bottomrule
\end{tabular}
\vspace{1.2em}

\begin{tabular}{l@{\hspace{5em}}ccc@{\hspace{5em}}ccc@{\hspace{5em}}ccc}
\toprule
\multicolumn{10}{c}{\textbf{HELILPR}} \\
\midrule
\textbf{Method}
& \multicolumn{3}{c@{\hspace{6em}}}{\textbf{RB01}} & 
\multicolumn{3}{c@{\hspace{6em}}}{\textbf{RB02}} & 
\multicolumn{3}{c}{\textbf{RB03}} \\
\midrule
& \textbf{R@1} & \textbf{F1} & \textbf{AUC}
& \textbf{R@1} & \textbf{F1} & \textbf{AUC}
& \textbf{R@1} & \textbf{F1} & \textbf{AUC} \\
\midrule
\multicolumn{10}{c}{\textbf{Netvlad}} \\
\midrule
BEV   & 0.836 & 0.928 & 0.977 & 0.537 & 0.712 & 0.772 & 0.603 & 0.866 & 0.947 \\
Polar & 0.898 & 0.941 & 0.979 & 0.698 & 0.830 & 0.828 & 0.726 & 0.857 & 0.926 \\
Range & 0.743 & 0.936 & 0.982 & 0.161 & 0.781 & 0.802 & 0.464 & 0.925 & 0.972 \\
Front & 0.315 & 0.542 & 0.262 & 0.103 & 0.988 & 0.975 & 0.379 & 0.846 & 0.899 \\
\midrule
\multicolumn{10}{c}{\textbf{MS}} \\
\midrule
BEV   & 0.739 & 0.980 & 0.993 & 0.169 & 0.776 & 0.802 & 0.425 & 0.922 & 0.970 \\
Polar & 0.753 & 0.987 & 0.997 & 0.131 & 0.905 & 0.899 & 0.463 & 0.946 & 0.983 \\
Range & 0.725 & 0.949 & 0.985 & 0.145 & 0.818 & 0.878 & 0.438 & 0.911 & 0.968 \\
Front & 0.322 & 0.529 & 0.268 & 0.101 & 0.975 & 0.979 & 0.361 & 0.819 & 0.876 \\
\bottomrule
\end{tabular}

\label{table:result3}
\end{table*}

\begin{table}[htbp]
\caption{Results of the models for the WAREHOUSE dataset. For each sequence, R, MF1, and AUC are reported.}
\centering
\fontsize{10}{11}\selectfont

\begin{tabular}{l@{\hspace{2em}}ccc}
\toprule
\multicolumn{4}{c}{\textbf{WAREHOUSE}} \\
\midrule
& \textbf{S1}\\
\midrule
\textbf{Method}
& \textbf{R@1} & \textbf{F1} & \textbf{AUC}\\
\midrule
\multicolumn{4}{c}{\textbf{Netvlad}} \\
\midrule
BEV   & 0.772 & 0.879 & 0.948  \\
Polar & 0.808 & 0.906 & 0.962  \\
Range & 0.805 & 0.889 & 0.946  \\
Front & 0.803 & 0.895 & 0.959  \\
\midrule
\multicolumn{4}{c}{\textbf{MS}} \\
\midrule
BEV   & 0.713 & 0.835 & 0.902  \\
Polar & 0.834 & 0.930 & 0.968  \\
Range & 0.771 & 0.870 & 0.949  \\
Front & 0.718 & 0.844 & 0.934  \\
\bottomrule
\end{tabular}

\label{table:result1}
\end{table}

\section{Experiments}
In this section, we provide information about the training and evaluation step. 
We use Dinov3 ViT-B/16 distilled with 86M parameters, and the projection head as described in Section \ref{sec:Methods}.
\subsection{Training}

We follow a reproducible, efficiency-aware training protocol designed to produce robust, viewpoint-invariant place descriptors while keeping fine-tuning computationally tractable. The approach combines a conservative adaptation of the pretrained backbone with aggressive, task-specific training of the projection head, uses pose-aware sampling to construct informative triplets, and applies periodic hard-negative mining to focus learning on difficult distinctions.

Because the pretrained backbone (DINOv3) and the projection head (NetVLAD) have different roles and parameter scales, we apply two distinct optimisation strategies. The backbone is fine-tuned using Low-Rank Adaptation (LoRA) \cite{hu2022lora,yang2024low}, which reduces the number of trainable parameters from 84M to 12M by learning low-rank updates; this enables conservative adaptation with less risk of overfitting. LoRA updates are trained with a learning rate of \(1\times 10^{-5}\) and weight decay \(0.04\). The NetVLAD head is trained fully (all head parameters are optimised) with a larger learning rate of \(1\times 10^{-3}\) and weight decay \(0.001\), allowing rapid, task-specific learning of the aggregation layer. The entire optimisation process is executed using the AdamW optimizer, an advanced variation of Adam that incorporates decoupled weight decay. Training optimises a triplet-loss: each tuple contains an anchor image, a positive (another image of the same place), and a negative (an image of a different place) \cite{park2024evaluation}. This objective explicitly encourages embeddings of the same place to be close and embeddings of different places to be far apart. We train on the KITTI \texttt{00} sequence (approximately 3000 frames) for 50 epochs, using in-plane random rotations of the anchor sampled uniformly in \([0,2\pi)\) to improve robustness to viewpoint changes. To prioritise difficult examples, we perform hard mining every two epochs. Concretely, after periodic evaluation on the training pool we identify negatives that the model currently confuses with positives (false negatives with small distance to the anchor) and add them to subsequent training batches. This focused curriculum improves discrimination and speeds convergence. Training tuples are constructed using pose information to produce informative positives and negatives: positives satisfy \(\|p_q-p_p\|_2 < 5\ \mathrm{m}\) (default) and negatives satisfy \(\|p_q-p_n\|_2 > 10\ \mathrm{m}\) (default). We typically sample \(K=10\) negatives per query and optionally use a cached feature bank to efficiently retrieve hard negatives for mining. Training runs on an NVIDIA Tesla V100-PCIE-16GB GPU. The combination of LoRA for the backbone and full training for the head balances computational cost and adaptability: LoRA keeps backbone updates compact and stable, while the fully trained NetVLAD head can quickly learn the aggregation behaviour required for place recognition. Fixed seeds, pose-aware sampling, and routine hard mining together make our experiments robust and comparable across runs and model variants. To ensure repeatability of results, we fix the random seed for NumPy and PyTorch (both CPU and CUDA). This reduces run-to-run variance and makes experimental comparisons reliable.

\subsection{Testing}\label{subsection:Testing}

For every dataset and for each sequence within it retrieval can be evaluated under three complementary test regimes. 

In all cases the evaluation uses the same core procedure: the model computes all global descriptors \(G=\{g_i\}_{i=1}^N\), the database vectors are indexed with FAISS (\texttt{IndexFlatL2}), and for each query the index returns the top-1 candidate \(\hat j_q\) and its L2 distance \(d_q=\|\tilde g_q-\tilde g_{\hat j_q}\|_2\). The chosen threshold for labeling a frame as revisited is $\tau = 5$. Queries that have no ground-truth positives (by the pose-based threshold) are excluded from the primary metrics; the reported accuracy, therefore, quantifies discriminability only where a revisited place exists.

\paragraph{Intra-sequence}  
This test evaluates retrieval under short- to medium-term viewpoint variation and repeated structure inside the same traversal. For a sequence we split frames at index \(N_{\mathcal D}\) into a \emph{database} \(\mathcal D=\{1,\dots,N_{\mathcal D}\}\) and \emph{queries} \(\mathcal Q=\{N_{\mathcal D}+1,\dots,N\}\). To avoid trivial temporal matches, the code applies an explicit temporal offset (200 frames by default) so that only temporally separated revisits are considered positives. For each query \(q\in\mathcal Q\) the FAISS top-1 \(\hat j_q\) is tested against the pose-derived positive set and aggregated into Recall@1, and the top-1 distances \(d_q\) are used to derive a precision–recall curve, max-F\(_1\) and PR AUC. 

\paragraph{Inter-sequence (same-trajectory, different-times).}  
This regime measures robustness to long-term, seasonal or illumination-driven appearance change by using one entire sequence as the database and other temporally-separated traversals of the same route as queries. Concretely, for a chosen database sequence we build a FAISS index on its vectors and run top-1 search for all frames in each other sequence that we want to evaluate as query set. This exposes the descriptor’s ability to match places across days/weeks and to tolerate large scene change while maintaining low false-positive rates.

\paragraph{Time Window}
Time Windows regime conceptualizes the structure of loop closure algorithms, where the the navigation algorithm searches for loop closure only within past measures and within a certain uncertainty radius of past entries.
For a query \(\,g_t\), the database \(DB\) thus contains the measurements gathered in the preceding window of size \(w\),
\[
DB = \{g_{t-w-\delta},\dots,g_{t-\delta} \},
\]
where the fixed offset \(\delta\) prevents spurious matches arising from consecutive, overlapping observations. Choosing \(w\) and the offset trades off recall (larger \(w\) finds more revisits) against computational cost and the risk of incorrect matches due to accumulating odometry error.

\begin{figure*}[ht]
    \centering
    \includegraphics[width=\textwidth]{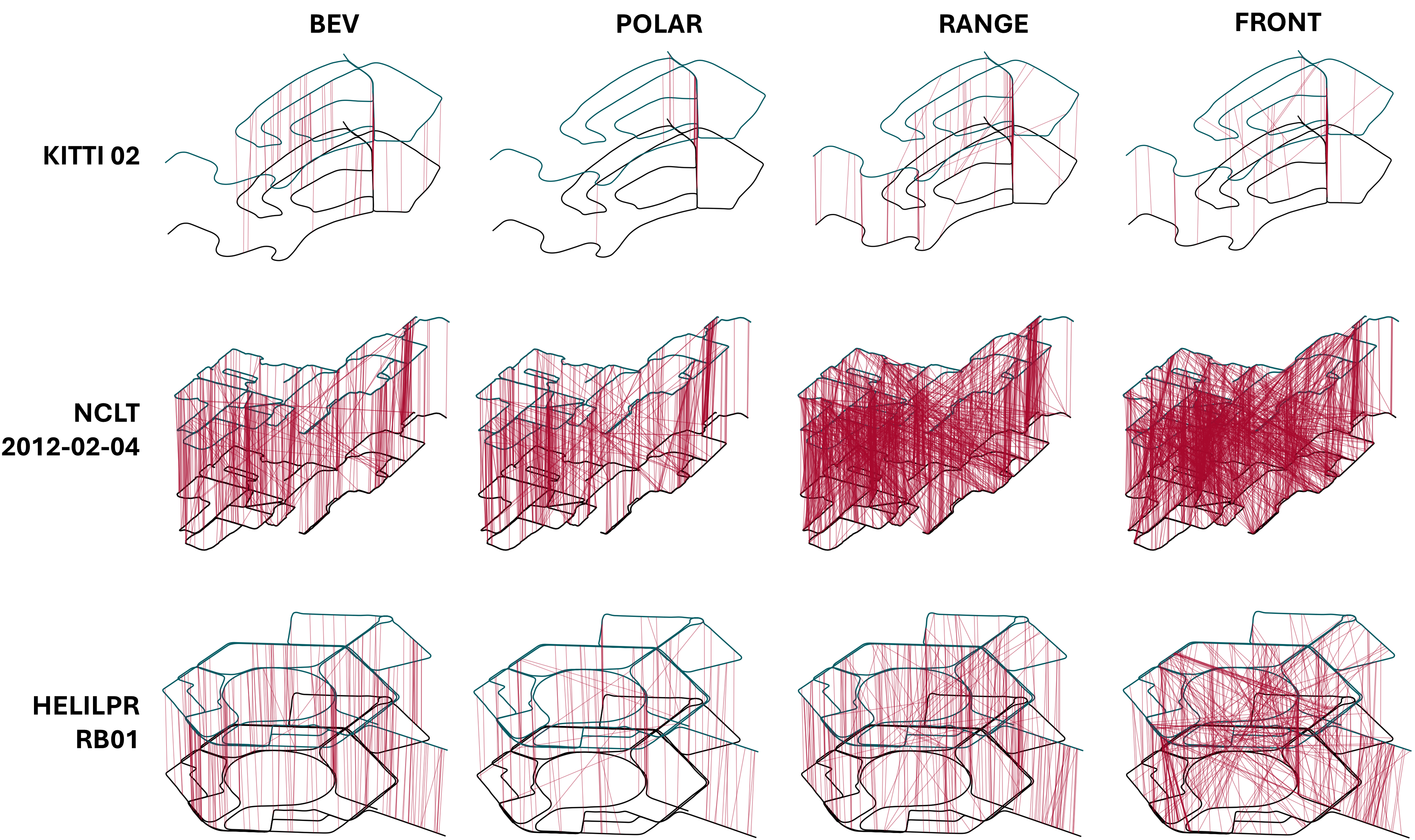}
    \caption{Visual concept of the place recognition capabilities across three different sequences (one for each public
    dataset)}
    \label{fig:map_recognition_cut}
\end{figure*}

\begin{figure*}[ht]
    \centering
    \includegraphics[width=\textwidth]{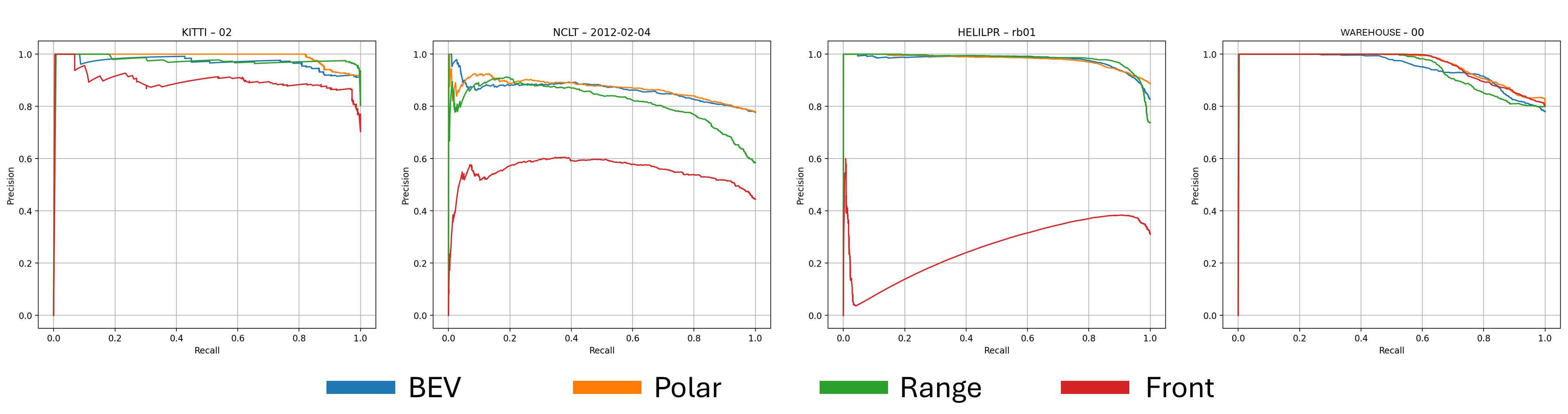}
    \caption{Precision recall for different representations and dataset}
    \label{fig:precision_recall}
\end{figure*}

\section{Results}
In Table \ref{table:result3} we list the evaluation results from the publicly available datasets. 

Across the three datasets, a coherent pattern emerges regarding the relative effectiveness of the considered representations for place recognition. The polar view consistently offers the strongest overall performance, showing superior robustness across varying sequences, sensing conditions, and temporal gaps. In KITTI, it achieves the highest values in nearly all metrics, slightly surpassing the BEV representation, while markedly outperforming both the range and front views, which exhibit weaker and less stable behaviour. A similar trend appears in NCLT, where the polar view maintains higher resilience to seasonal and long-term variations, again providing more reliable results than BEV and clearly exceeding the more fragile range and front representations, whose performance declines sharply under significant domain changes. In the HELILPR dataset, polar and BEV remain the two strongest contenders, with polar generally maintaining a slight edge in single-session scenarios, whereas range and especially front views continue to show considerable instability and lower discriminative capability. These observations indicate that the polar representation is the most consistently effective choice for robust place recognition, with BEV as a solid secondary option, while range and front views are limited in applicability due to their sensitivity to viewpoint variability and environmental changes. This result is confirmed in the WAREHOUSE dataset, presented in Table \ref{table:result1}. With this dataset, performance appears more uniform across all representations, a behaviour attributed to the feature-dense indoor environment and the continuous changes in scene appearance, which reduce the impact of viewpoint-dependent biases. Although the polar view still attains the highest overall scores—reaching an R@1 of $0.80$, F1 of $0.906$, and AUC of $0.962$ in the NetVLAD setting, and further improving to $0.834$, $0.930$, and $0.968$, respectively with the mean-standard deviation head, the remaining representations narrow the performance gap substantially. BEV, range, and front views cluster closely around similar recognition rates, each achieving R@1 values between $0.77$ and $0.80$ in the single-session experiment and maintaining comparable F1 and AUC scores. Even with mean-standard deviation setting, where polar preserves a modest advantage, the differences between representations remain smaller than in the outdoor datasets, reflecting a domain in which structural regularity and rich local texture diminish the relative influence of the chosen projection model.

The precision–recall curves (Figure \ref{fig:precision_recall} further reinforce the trends observed in the quantitative metrics, while also clarifying how each representation behaves across different operating regions. In KITTI, all representations except the front view maintain high precision over nearly the entire recall range, with the polar projection exhibiting the most stable curve and a slightly delayed drop-off, confirming its stronger discriminative power under urban driving conditions. The range and BEV curves remain competitive, though their precision oscillates more noticeably at intermediate recall, indicating a higher sensitivity to local structural variations. In contrast, the front view shows a substantial degradation in both early and mid-recall regions, reflecting its inherent vulnerability to viewpoint changes and partial occlusions.

A similar pattern emerges in NCLT, where the polar and BEV representations preserve higher precision as recall increases, although the overall curves shift downward due to the more challenging long-term appearance changes characteristic of the dataset. The range representation exhibits a pronounced decline in precision at moderate recall values, and the front view again collapses rapidly, confirming its lack of robustness in environments with strong seasonal variability. For HELILPR, the polar, BEV, and range projections produce tightly clustered curves with high precision across most of the recall spectrum, reflecting the more geometrically stable conditions of the dataset; nonetheless, polar retains a slight advantage near the end of the curve. The front view, by contrast, shows a markedly different trajectory, with extremely low precision at low recall and a slow monotonic increase, underscoring its inability to provide reliable matches under aerial or semi-aerial viewpoints.

The WAREHOUSE sequence presents a more balanced scenario, where BEV, polar, and range curves overlap for most of the recall range, maintaining high precision before gradually decaying toward the upper end. Here the differences among these three projections are minor, consistent with a dataset where viewpoint variability and environmental clutter are less pronounced. The front view converges with the other methods only at early recall but falls more sharply later, mirroring its tendency to overfit specific perspectives rather than general scene structure. Overall, these curves highlight that while BEV, polar, and range each achieve strong performance under certain conditions, the polar projection consistently yields the most stable precision–recall behaviour across datasets with diverse appearance changes, whereas the front view remains the least reliable due to its pronounced susceptibility to perspective-dependent artifacts.

In \ref{fig:map_recognition_cut}, we show a visual concept of the place recognition capabilities across three different sequences (one for each public dataset). The procedure involves two steps: first, computing all map descriptors for the established trajectory; second, for every query point in the trajectory, identifying the two nearest neighbors within the set of pre-computed map descriptors. A place is defined as correctly recognized if the second-closest neighbor is located within a Euclidean distance of $5m$ from the query point's ground truth coordinates, assuming the first-closest neighbor is the query point itself. The recognition statistics derived from this process confirm the quantitative trends previously observed and presented in Table \ref{table:result3}.

 \section{Discussion}

The results in Table \ref{table:result3} reveal a clear and consistent advantage for the polar representation: across diverse outdoor benchmarks and in the custom WAREHOUSE sequence the polar projection yields the most stable and highest retrieval scores. Notably, the polar projection demonstrates a consistently stronger and more robust behaviour than the competing representations across sensing conditions and temporal gaps.

Two complementary pieces of evidence support this claim. First, the scalar metrics (R@1, max-F\(_1\), PR-AUC) show polar either matching or exceeding BEV and substantially outperforming range and front projections in KITTI, NCLT and HELILPR. Second, the precision–recall curves (Fig.~\ref{fig:precision_recall}) demonstrate that polar maintains higher precision over a wider recall range and exhibits a gentler drop-off than competing representations, i.e. it is not only more accurate at a particular operating point but also better calibrated as a confidence signal.

Why does polar work better? Intuitively, polar coordinates respect the native sampling geometry of spinning LiDAR: points are organized by bearing and range around the sensor, so a polar projection preserves neighbourhood relations and angular continuity that are meaningful for loop closure. This representation reduces sensitivity to lateral viewpoint shifts and to partial occlusions: a rotated revisit often appears as a circular shift in the polar domain and therefore produces more consistent patch-level correspondences than perspective (front) or heavily warped range images. BEV remains a strong secondary choice because it encodes global scene layout and large-scale structure well, but it is somewhat more sensitive to vertical structure and viewpoint-dependent occlusions than polar.
Range and front projections are the weakest and most variable in outdoor datasets because they amplify perspective distortions and occlusions: the front view collapses 3D structure into a single azimuthal slice and is therefore brittle to viewpoint and elevation changes, while the range image suffers from upsampling artifacts and distortion near the sensor. These effects explain the sharp precision drops and unstable mid-recall behaviour seen in KITTI and NCLT.
The WAREHOUSE results (Table \ref{table:result1}) clarify the applicability limits and the practical feasibility of the range and front views. In this dense, indoor–outdoor, feature-rich environment the performance gap between polar, BEV and range narrows substantially and all three projections achieve similar single-session recognition rates (R@1 in the \$0.77\text{--}0.80\$ range). Two factors explain this: (i) the scene contains abundant, repeatable local structure (shelves, walls, machinery) that produces rich LiDAR returns even in perspective or range encodings; and (ii) viewpoint variability is often constrained by aisle-like trajectories and repeated traversals, which reduces the distortions that typically break front and range representations in unstructured outdoor scenes. Consequently, range and (to a lesser extent) front projections become feasible options in warehouse-like settings where dense geometry and constrained motion yield stable, discriminative patterns in those domains.
Putting these observations together yields practical guidance. For general outdoor deployment and long-term localisation we recommend polar as the default projection due to its superior and consistent robustness; use BEV as a reliable fallback when global layout cues are paramount. Range and front encodings should be considered primarily for structured, feature-dense indoor environments (e.g., warehouses, corridors) or as part of a multi-projection ensemble where their complementary signal can help in specific scenarios — but operators should be aware of their heightened sensitivity to viewpoint and occlusion in large, open, or seasonally varying scenes. Finally, although NetVLAD generally produces larger and more discriminative global descriptors, the WAREHOUSE experiments show that representation choice and dataset geometry interact with the projection head; this suggests that modest head-specific tuning (or simple ensembling across projections) can further improve real-world robustness without changing the core sensor pipeline.

\section{Conclusions}

Our study set out with a single, focused question: which 2-D projection of a LiDAR scan produces the most discriminative and robust place descriptor when processed by a modern vision foundation backbone? The empirical answer is clear and practically useful: a sensor-centric polar projection yields the most consistent and well-calibrated descriptors across multiple outdoor benchmarks and a challenging, dynamic warehouse sequence, with BEV as a reliable secondary choice. Crucially, this finding enables the leverage of off-the-shelf vision foundation models for LiDAR-based place recognition by merely requiring lightweight finetuning to adapt their inherent generalization capabilities to the Visual Place Recognition (VPR) task: an appropriate projection preserves the geometric cues that these backbones need and lets compact aggregation heads produce strong retrieval performance. The range and front encodings, while generally less robust in open, long-term outdoor settings, proved entirely feasible in the feature-dense, constrained motions of the warehouse domain (we consider it a reminder that dataset geometry and deployment context should guide projection choice). We also show that projection and aggregation head interact: head design (e.g., NetVLAD vs simple statistics) matters and can amplify—or mitigate—representation strengths. In short, the paper gives a concrete, experimentally validated recommendation for practitioners and lays out several low-risk paths for improvement (projection–head co-design, lightweight ensembling, and tighter odometry integration) rather than calling for a wholesale architectural rework.

\bibliography{bibliography}
\bibliographystyle{IEEEtran}

\newpage

 




\vfill

\end{document}